\documentclass[10pt,twocolumn,letterpaper]{article}

\usepackage{wacv}
\usepackage{times}
\usepackage{epsfig}
\usepackage{graphicx}
\usepackage{amsmath}
\usepackage{amssymb}
\usepackage{booktabs}
\usepackage{multirow}
\usepackage{makecell}
\usepackage{subcaption}
\usepackage{stfloats}
\usepackage{balance}
\usepackage{enumitem}
\usepackage{stackengine}
\usepackage{textcomp}
\usepackage{caption}
\usepackage{float}
\usepackage{tablefootnote}
\usepackage{threeparttable}
\usepackage{capt-of,etoolbox}
\usepackage{array}
\usepackage{enumitem}
\usepackage{xcolor}

\setlist{nosep} 

\newcommand{\Lagr}{\mathcal{L}}

\def\wacvPaperID{277} 

\wacvfinalcopy 

\ifwacvfinal
\fi

\ifwacvfinal
\usepackage[breaklinks=true,bookmarks=false]{hyperref}
\else
\usepackage[pagebackref=true,breaklinks=true,colorlinks,bookmarks=false]{hyperref}
\fi

\ifwacvfinal
\else

\fi

\makeatletter
\apptocmd\@maketitle{{\myfigure{}\par}}{}{}
\makeatother

\begin{document}

\title{Domain-Aware Unsupervised Hyperspectral Reconstruction\\for Aerial Image Dehazing}
\author{
Aditya Mehta\textsuperscript{1}\hspace{1em} Harsh Sinha\textsuperscript{1} \hspace{1em} Murari Mandal\textsuperscript{2} \hspace{1em} Pratik Narang\textsuperscript{1}\\
\textsuperscript{1} Department of CSIS, BITS Pilani, India\\
\textsuperscript{2} Department of CSE, IIIT Kota, India \\
{\tt\small \{f2015808,h20130838,pratik.narang\}@pilani.bits-pilani.ac.in},
{\tt\small murarimandal.cv@gmail.com}

}



\newcommand\myfigure{%
\centering
    \includegraphics[width=.8\textwidth]{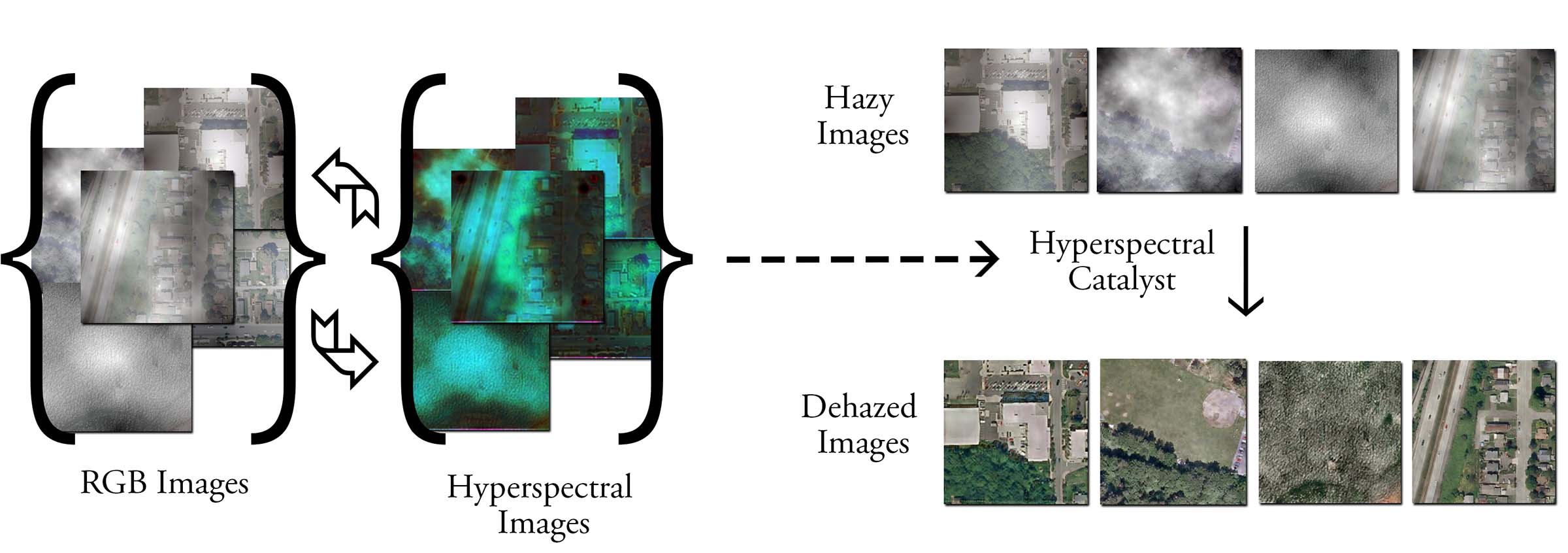}
\captionof{figure}{Left: Unsupervised hyperspectral reconstruction for hazy aerial images. Right: Once trained, features derived from reconstructed hyperspectral images serve as a catalyst to dehaze aerial images}
\label{fig:teaser}
\medskip
}

\maketitle


\begin{abstract}
Haze removal in aerial images is a challenging problem due to considerable variation in spatial details and varying contrast. Changes in particulate matter density often lead to degradation in visibility. Therefore, several approaches utilize multi-spectral data as auxiliary information for haze removal. In this paper, we propose SkyGAN for haze removal in aerial images. SkyGAN consists of 1) a domain-aware hazy-to-hyperspectral (H2H) module, and 2) a conditional GAN (cGAN) based multi-cue image-to-image translation module (I2I) for dehazing. The proposed H2H module reconstructs several visual bands from RGB images in an unsupervised manner, which overcomes the lack of hazy hyperspectral aerial image datasets. The module utilizes task supervision and domain adaptation in order to create a ``hyperspectral catalyst'' for image dehazing. The I2I module uses the hyperspectral catalyst along with a 12-channel multi-cue input and performs effective image dehazing by utilizing the entire visual spectrum. In addition, this work introduces a new dataset, called Hazy Aerial-Image (HAI) dataset, that contains more than 65,000 pairs of hazy and ground truth aerial images with realistic, non-homogeneous haze of varying density. The performance of SkyGAN is evaluated on the recent SateHaze1k dataset as well as the HAI dataset. We also present a comprehensive evaluation of HAI dataset with a representative set of state-of-the-art techniques in terms of PSNR and SSIM.
\end{abstract}

\section{Introduction}\label{sec:intro}

Aerial imagery refers to photographs taken from aircraft such as helicopters and UAVs. Such images are advantageous as they have rich information content. Hence, they have been widely utilized in various fields such as remote sensing~\cite{lillesand2015remote}, earth sciences~\cite{mustard1999spectral}, agriculture~\cite{haboudane2004hyperspectral} and geology~\cite{cloutis1996review}. The aerial images and video data facilitate numerous applications such as
aerial surveillance~\cite{mandal2020mor,mandal2019avdnet,mandal2019sssdet}, search and rescue, event recognition~\cite{mou2020era}, urban and rural scene understanding~\cite{xia2018dota,garg2020isdnet}. As the aerial images are perceived from a considerable distance, these images often suffer from low visibility, color shift, and blurriness due to changes in the atmospheric path. Such atmospheric effects, especially non-homogeneous clouds, fog, and haze, degrade the quality of input images. Therefore, there is a need to address visibility improvement before aerial images can be used in aerial vision-based systems.\par

Researchers have used several different techniques to remove haze in aerial images. Zhang et al.~\cite{zhang2003quantitative} used a correction technique using a correlation between the high-frequency and the low-frequency color bands. Such methods quantify the amount of haze using different spectral bands in the visible region. Liu et al.~\cite{liu2011haze} present a virtual cloud point method based on relative haze thickness for haze removal. Long et al.~\cite{long2013single} removed haze in remote sensing images by adapting the DCP proposed by He et al.~\cite{he2010single} for natural scene images. Similarly, Shen et al.\ ~\cite{shen2020spatial} suggested a gradient-based spectral adaptive approach to exploit the wavelength-dependent transmission information. Makarau et al.~\cite{ makarau2016combined} have used NIR and multispectral images to exploit the multi-sensor information for accurate haze removal. However, the use of hyperspectral images (HSI) in deep learning-based haze removal has been limited. HSI consists of spectral reflectance information from a substantial number of wavebands. The high cost of HSI acquisition devices and the lack of large-scale HSI datasets are some of the impediments in wide-scale acceptance of HSI in general computer vision. Acquisition of full spectral signatures using a HSI camera is a costly endeavor, not only in terms of hardware but it also hampers the viability of UAVs and other radio-controlled aerial vehicles.\par
\begin{figure*}[t!]
\centering
    \begin{subfigure}[t]{0.09\textwidth}
        \includegraphics[width=\textwidth]{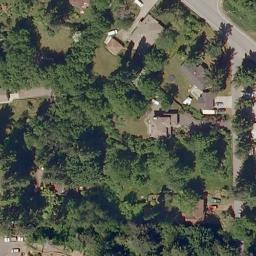}
        \includegraphics[width=\textwidth]{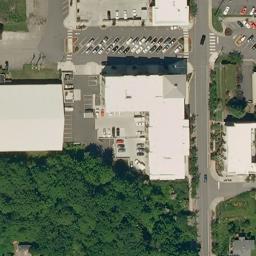}
        \includegraphics[width=\textwidth]{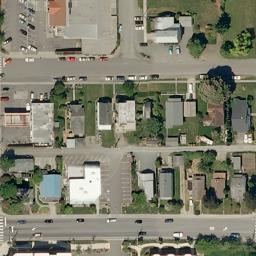}
    \end{subfigure}
        \begin{subfigure}[t]{0.09\textwidth}
        \includegraphics[width=\textwidth]{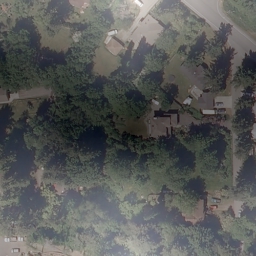}
        \includegraphics[width=\textwidth]{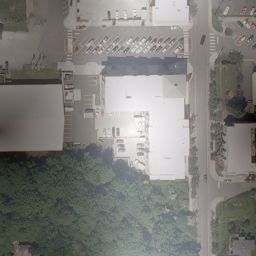}
        \includegraphics[width=\textwidth]{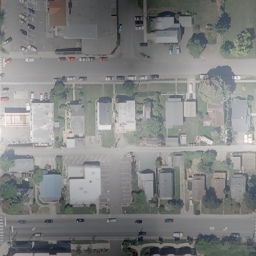}
    \end{subfigure}
    ~
    \begin{subfigure}[t]{0.09\textwidth}
        \includegraphics[width=\textwidth]{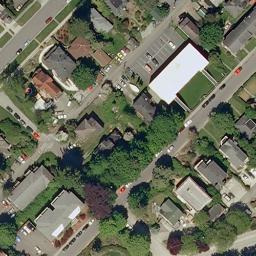}
        \includegraphics[width=\textwidth]{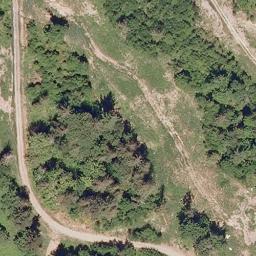}
        \includegraphics[width=\textwidth]{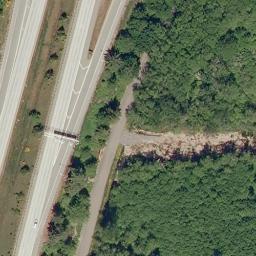}
    \end{subfigure}
        \begin{subfigure}[t]{0.09\textwidth}
        \includegraphics[width=\textwidth]{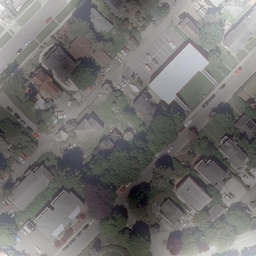}
        \includegraphics[width=\textwidth]{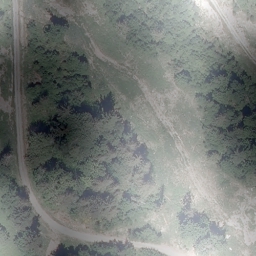}
        \includegraphics[width=\textwidth]{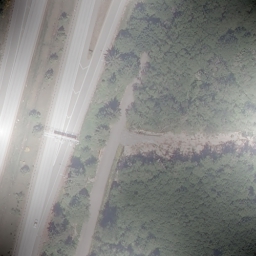}
    \end{subfigure}
    ~
    \begin{subfigure}[t]{0.09\textwidth}
        \includegraphics[width=\textwidth]{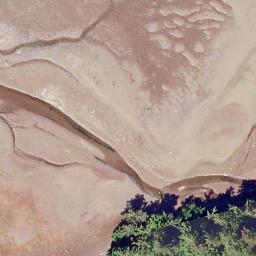}
        \includegraphics[width=\textwidth]{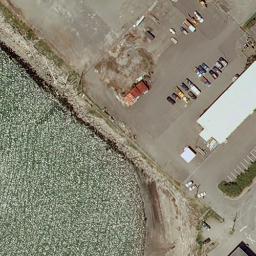}
        \includegraphics[width=\textwidth]{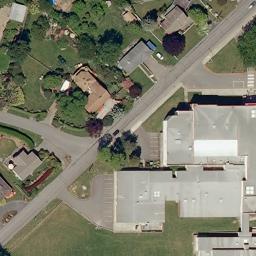}
    \end{subfigure}
    \begin{subfigure}[t]{0.09\textwidth}
        \includegraphics[width=\textwidth]{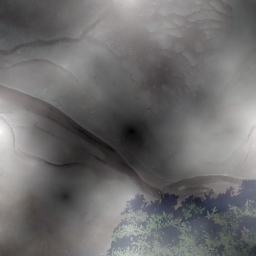}
        \includegraphics[width=\textwidth]{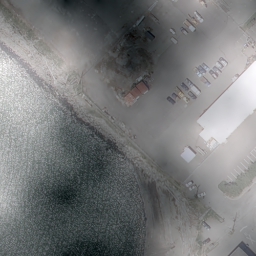}
        \includegraphics[width=\textwidth]{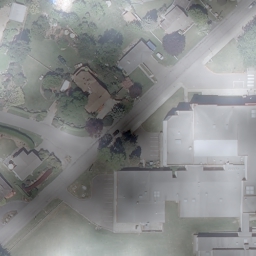}
    \end{subfigure}
    ~
    \begin{subfigure}[t]{0.09\textwidth}
        \includegraphics[width=\textwidth]{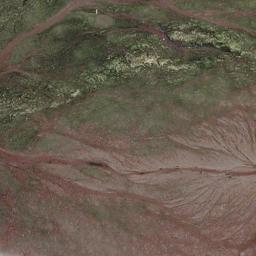}
        \includegraphics[width=\textwidth]{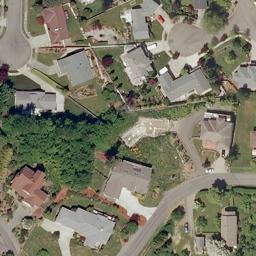}
        \includegraphics[width=\textwidth]{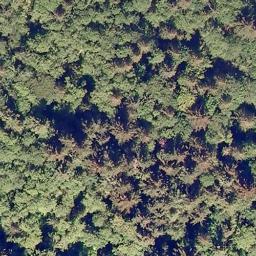}
    \end{subfigure}
    \begin{subfigure}[t]{0.09\textwidth}
        \includegraphics[width=\textwidth]{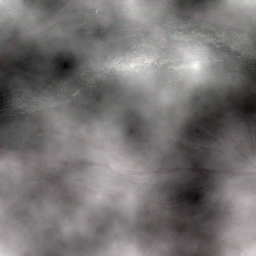}
        \includegraphics[width=\textwidth]{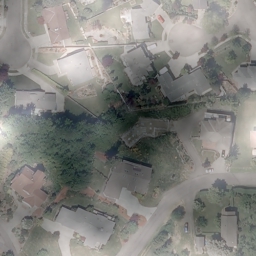}
        \includegraphics[width=\textwidth]{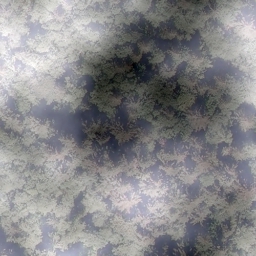}
    \end{subfigure}
    ~
    \begin{subfigure}[t]{0.09\textwidth}
        \includegraphics[width=\textwidth]{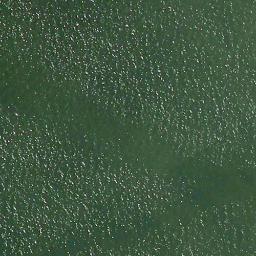}
        \includegraphics[width=\textwidth]{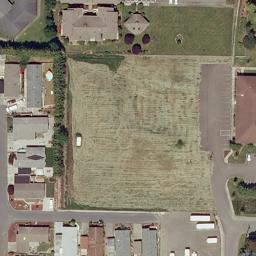}
        \includegraphics[width=\textwidth]{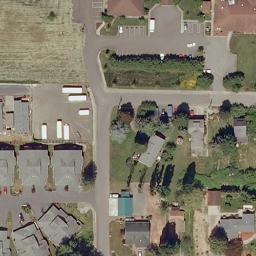}
    \end{subfigure}
    \begin{subfigure}[t]{0.09\textwidth}
        \includegraphics[width=\textwidth]{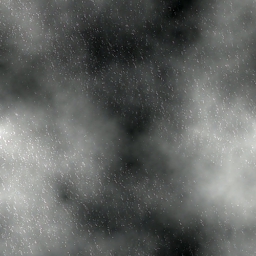}
        \includegraphics[width=\textwidth]{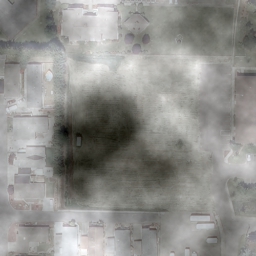}
        \includegraphics[width=\textwidth]{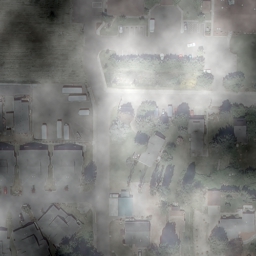}
    \end{subfigure}
    
    \label{fig:HAI_show}
\caption{Illustration of corresponding clean and hazy images from the proposed HAI dataset which consists of 5 different degrees of haze, represented in different columns.}
\end{figure*}

In this work, the input hazy RGB image is used to reconstruct a domain-aware HSI using a modified cycle-consistent framework. The proposed framework is a low-cost and efficient approach to generate HSI in an unsupervised manner. Further, we present the practical applicability of reconstructed HSI for aerial image dehazing. The key motivation to use HSI for dehazing is to utilize the entire visual spectrum which provides rich information content in comparison to three primary bands: red, green and blue. Since the perception of an object depends on its nature to absorb different wavelengths, this work proposes to enhance the visual contrast of an image by acquiring information from different spectral regions. 

The key contributions of this work are as follows:
\begin{enumerate}[itemsep=0em]
    \item A novel GAN framework named SkyGAN is presented which incorporates HSI guidance in image-to-image translation network for image dehazing. To the best of our knowledge, it is the first attempt to use HSI in a GAN framework for aerial image dehazing.
    \item SkyGAN reconstructs HSI from RGB images in an unpaired manner, which addresses the lack of hazy HSI datasets.
    \item The architecture utilizes task supervision in conjunction with learning closed set domain adaptation by combining two techniques,  adversarial distribution discrepancy alignment~\cite{ganin2014unsupervised} and cycle-consistency constraint~\cite{zhu2017unpaired}, to reconstruct domain-aware HSI from a set of natural scene HSI.
    \item A residual network is used to alleviate data distribution discrepancy from reconstructed HSI, thereby generating a hyperspectral catalyst (HSC) which is fed along with a 12-channel multi-cue input to the proposed conditional GAN (cGAN) for haze removal.
    \item Further, we introduce a large-scale hazy aerial image dataset HAI, consisting of over 65 thousand pairs of hazy and ground truth (haze-free) aerial images. The images in HAI contain realistic, non-homogeneous haze of varying density. 
    \item The proposed SkyGAN outperforms the existing state-of-the-art approaches on the benchmark SateHaze1k \cite{huang2020single} and the new large-scale HAI dataset. We also present an extensive ablation study of the different components of SkyGAN.
\end{enumerate}

\begin{figure*}
    \centering
    \includegraphics[width=\linewidth]{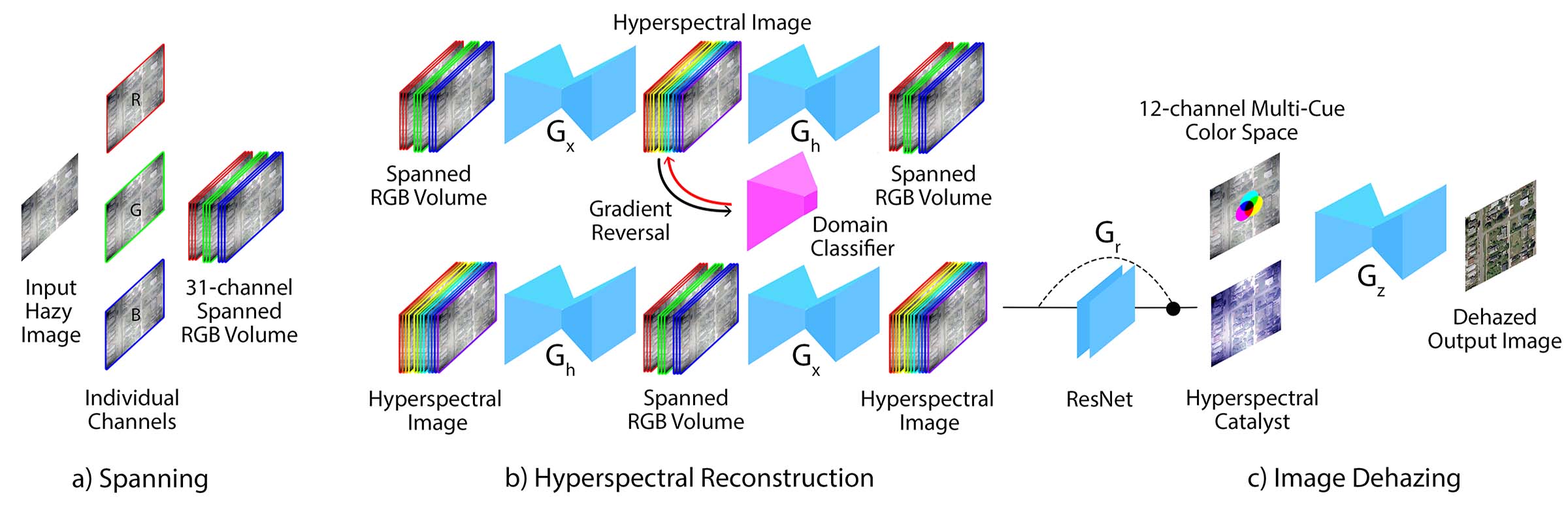}
    \caption{Architecture for the proposed SkyGAN. The corresponding discriminators $D_x, D_h$ and $D_z$ for generators $G_x, G_h$ and $G_z$ respectively are not displayed for brevity of the diagram.}
    \label{fig:my_label}
\end{figure*}


\section{Related Work}\label{sec:relwork}

The existing dehazing approaches can be categorized into radiative transfer (RT) models which require atmosphere ambience related parametric information, and statistical information (SI) methods which rely on image characteristics for haze removal. 
RT models build upon some assumption or a physical prior such as contrast~\cite{tan2008visibility}, dark channel~\cite{he2010single}, color-attenuation~\cite{zhu2015fast}, and non-local prior~\cite{berman2016non}. On the other hand, there have been several SI approaches that typically learn the transmission map~\cite{cai2016dehazenet} or a direct end-to-end mapping between hazy and clean images~\cite{li2017aod,mehra2020reviewnet,qu2019enhancedpix2pix,ancuti2020ntire}. With advancements in deep learning, researchers have employed GANs for image dehazing task by learning the transmission map~\cite{zhang2018densely,engin2018cycle,qu2019enhanced,dudhane2019ri}.
Researchers have also used hyperspectral images with GANs for super-resolution \cite{gwn2019generative}, classification \cite{zhu2018generative} and natural scene dehazing \cite{mehta2020hidegan}.   
\par

The existing techniques use standard RGB images for dehazing. HSI can be utilized to enhance segmentation and classification by using rich information from the entire visual spectrum. The use of HSI has been restricted primarily due to its high acquisition cost. Therefore researchers have developed techniques such as multiplexed illumination~\cite{goel2015hypercam} and sparse dictionary prior~\cite{arad2016sparse} to reconstruct HSI from RGB images which alleviate the need for costly acquisition devices. Despite the advantages of using HSI, it has not been accepted in general vision tasks due to a lack of large scale HSI datasets. Thus, we propose an unsupervised approach for HSI reconstruction from RGB using domain adaptation. 

\section{Proposed Method}\label{sec:proposed}
The proposed approach benefits from rich information obtained from different bands in HSI reconstructed from standard RGB images. HSI reconstruction have been achieved via prior-based techniques. However, such an approach is infeasible as the existing aerial image datasets are restricted to infrared, synthetic aperture radar or multispectral images. Therefore, we adopt an approach built upon unsupervised domain adaptation to reconstruct haze-invariant HSI for aerial images.

Given a 3-channel image, $\mathbf{a}_i \in {\rm I\!R}^3 $ as input, we solve the problem of hallucinating a plausible spectral signature of the input image. As the problem is under-constrained, we re-formulate the reconstruction task as a spectral reflectance interpolation problem. To achieve this, the individual channels of the input images are spanned to form a 31-channel input. Let an input image $\mathbf{a_i} \in {\rm I\!R}^{h\times w \times 3} $. The individual channels in $\mathbf{a_i}$ are spanned to form a spanned image matrix $\mathbf{x}_{i} \in {\rm I\!R}^{h\times w \times 31} $ where $h, w$ denote the image dimensions. The spanned image $\mathbf{x}_{i}$ is used in turn for generating the HSI.


\subsection{Hazy-to-Hyperspectral Reconstruction (H2H)}\label{subsec:h2h}

Given $\{h_i\}_{i=1}^{n_s}$ of HSI sampled from the source domain of natural scenes ${\cal{H}}$, and $\{\mathbf{x}_j\}_{j=1}^{{n}_t}$ of standard RGB images sampled from the target domain of aerial images ${\cal{X}}$, the objective of proposed unsupervised domain adaptation framework is to learn how to translate between these domains without paired image examples. The algorithm is based on the assumption that there is a rich underlying HSI dictionary that can be used to reconstruct novel images from the distribution of hyperspectral signatures in natural images. The proposed methodology makes use of several guiding principles such as domain adaptation, task supervision and domain classifier.

\paragraph{Unsupervised Domain Adaptation.} First of all, we lack supervision in the form of paired examples. To address the problem, the proposed algorithm exploits supervision at the level of sets. Given a set of HSI and a set of standard RGB images, we train a mapping $G_x(\cdot) : \cal{X} \to \cal{H}$ such that the spectral response of the output $\hat{h} = G_x({x})$, $x\in \cal{X}$ is indistinguishable from spectral response of $h \in \cal{H}$ by an adversary trained to classify $\hat{h}$ apart from $h$. In addition, we also train a mapping $G_h(\cdot) : \cal{H} \to \cal{X}$ which introduces cycle-consistency, thus adding more structure in the proposed algorithm. 
{
In context to adversarial losses, the proposed network minimizes the following objective:

\begin{align}
     \begin{split}
    \Lagr_{x}(G_x,D_x, x, h) &= \mathbb{E}_{x\sim \cal{X}}[log(1-D_x(G_x(x)))] \\
&+\mathbb{E}_{h\sim \cal{H}}[log(D_x(h))] \\
    \end{split}
    \\
    \begin{split}
    \Lagr_{h}(G_h,D_h, h, x) &=  \mathbb{E}_{h\sim \cal{H}}[log(1-D_h(G_h(h)))]
\\ &+\mathbb{E}_{x\sim \cal{X}}[log(D_h(x))] \\
    \end{split}
    \\
    \begin{split}
    \Lagr_{GAN} &= \Lagr_{x} + \Lagr_{h} \\
    \end{split}
\end{align}
}
\paragraph{Task Supervision.} {
If we consider 3-channel RGB images in comparison to 31-channel hyperspectral images, hallucinating a plausible spectral response seems daunting. Comparatively, the RGB image has lost 90\% of information. Although, the semantics of images and the textural cues provide ample guidance to reconstruct hyperspectral counterparts they are not sufficient to retrieve the actual ground truth. Thus, we aim at learning a \textit{plausible} HSI image for a given input RGB image by modelling the latent hyperspectral space for image derived from a set of hyperspectral images.

Due to lack of hyperspectral images for dehazing, we adopt an unpaired approach for hyperspectral reconstruction. For each iteration, an input $x$ is mapped to $h$ by means of a stochastic mapping $x\sim{p_{\cal{H, X}}(\hat{h}|x)}$. The objective is to learn a plausible spectral response of given RGB image by learning a latent hyperspectral space. Thus, it can be inferred that the $G_x(\cdot)$ can be learned efficiently if $G_x(\cdot)$ is stochastic in nature. This is required to build an output distribution over $\hat{h}$ that reconciles with original distribution $p_{\cal{H}}(h)$.
However,  with limited variation in small HSI datasets, it is problematic to learn an output distribution over $\hat{h}$ that matches its underlying hyperspectral space.}

In general, existing approaches address the issue of distributional discrepancy by a form of maximum-mean discrepancy. Rather than relying on a minimax optimization scheme for obtaining aerial HSI, we use our original dehazing task as a self-supervised auxiliary task~\cite{sun2019unsupervised}. Jointly training the HSI reconstruction task along with dehazing task supervision aligns the domains in a subspace specifically for image dehazing.
We modify the cycle-consistency loss \cite{zhou2016learning} to incorporate dehazing task supervision. The modified loss encourages $y \approx G_h(G_x(x))$ and  $h \approx G_x(G_h(h))$.

\begin{equation}
\label{eq1}
\begin{split}
\Lagr_{cyc} =& \Arrowvert y-G_h(G_x(x)) \Arrowvert_2^2 + \Arrowvert h-G_x(G_h(h)) \Arrowvert_2^2. 
\end{split}
\end{equation}
where $(x, y)$ refers to hazy and dehazed/clean RGB images respectively, and $h$ refers to HSI. 

\paragraph{Domain Classifier.} As the primary task of the framework is to produce natural dehazed images, we incorporate a domain classifer~\cite{ganin2014unsupervised} to learn a mapping $G_x(\cdot)$ which is inherently haze invariant. The motivation is to induce alignment between hazy and clean images by learning a domain classifier as an approximation to the total variation distance. This is achieved by jointly optimizing the algorithm with a domain classifier that discriminates between hazy and clear images during training. By maximizing the loss of domain classifier, the algorithm encourages haze-invariant HSI i.e.\ $\hat{h}$ in the course of parameter optimization such that the expected target risk ${\mathbb{E}}_{(\mathbf{x}^t)\sim \cal{X}}$ is low for loss function ${\cal{L}}_{cls}(\cdot)$,
\begin{equation}
\label{eq2}
\begin{split}
\Lagr_{cls} =& {\mathbb{E}}_{(\mathbf{x}^t)\sim \cal{X}}\Arrowvert h-G_x(G_h(h)) \Arrowvert_2^2 
\end{split}
\end{equation}
where $(x, h)$ refers to hazy RGB image and HSI respectively.

\paragraph{Implementation Details.} The generators in $G_x$, $G_h$ and $G_z$ adopt a U-Net with skip connections while PatchGAN is adopted for the corresponding discriminators. Although the RGB and HSI have considerable difference in the pixel domain, the underlying structure must remain intact. Thus, we use 
L2 cycle consistency losses and identity losses~\cite{zhu2017unpaired} to further improve the reconstructed HSI. Finally, a ResNet-based architecture is used to extract relevant features from reconstructed HSI, referred to as hyperspectral catalyst, which is further fed into a modified cGAN for efficient dehazing.

\subsection{Multi-Cue Image-to-Image Translation (I2I)}\label{subsec:i2i}
To achieve dehazed image outputs in an efficient manner, we use an enhanced cGAN which produces final dehazed output images by utilizing a 12-channel multi-cue color space image as input along with the hyperspectral catalyst. 

\paragraph{Multi-cue Color Space.} The multi-cue color space is constructed using different color models such as HSV, YCrCb, and LAB concatenated together with RGB which results in a 12-channel input image. 

The HSV color model is a cylindrical transformation of RGB Cartesian space in terms of hue, saturation, and luminance value. HSV provides a practical advantage as it separates saturation and luminance, and it has been used extensively in computer vision. Zhang et al.~\cite{zhang2018naturalness} argues that the HSV color space can be used to preserve hue~\cite{wan2015joint} and reduce computational complexity. The hypothesis is based on the fact that HSV specifically models visuals as perceived by human eye which helps in suppressing halo effects. 

The YCbCr represents the luminance, blue-difference, and red-difference chroma constituents, respectively. It provides a practical approximation for color processing and perceptual uniformity. The key motivation behind using a YCbCr color space is its robustness towards haze and color distortion~\cite{zhang2018naturalness,wan2015joint}. The use of multi-channel color space enhances the naturalness of the dehazed images~\cite{guo2010automatic,xie2010improved}. 

The LAB color space defines a color model that is more perceptually linear than any other color model approximating human perceptual vision. The advantage is especially significant in applications addressing visual degradation as it provides balance corrections to adjust lightness and contrast. Guo et al.~\cite{guo2010automatic} and Xie et al.~\cite{xie2010improved} used the Retinex algorithm on the luminance component of the color space to generate the transmission map from a single hazy image.

The proposed method allows the learning algorithm to use different color spaces simultaneously being pivoted on attaining optimal dehazed outputs.
\begin{table}
    \centering
    \begin{threeparttable}
    \begin{tabular}{c|l l}
    \toprule
    Method & \multicolumn{1}{c}{SSIM} & \multicolumn{1}{c}{PSNR} \\
    \midrule
    \midrule
    DCP~\cite{he2010single} & 0.5430  & 14.777 \\ 
    CAP~\cite{zhu2015fast} & 0.6342 & 10.582  \\ 
    AOD~\cite{li2017aod} 
    & 0.6499  & 10.227  \\
    GFN~\cite{ren2018gated} & 0.5340 & 16.787 \\
    CycleGAN~\cite{zhu2017unpaired} 
    & 0.5950 & 15.707  \\ 
    GridDehazeNet~\cite{liu2019griddehazenet} & \textbf{0.8915} & 22.543 \\
    \textbf{SkyGAN} & 0.8466 & \textbf{24.560} \\ 
    \bottomrule
    \end{tabular}
    \end{threeparttable}
    \caption{Comparative results on HAI dataset}
    \label{tab:my_label}
\end{table}

\paragraph{Hyperspectral Catalyst (HSC). } It is difficult for the learning algorithm to alleviate problems such as occlusion caused by haze in aerial images. The reconstructed HSI contains spectral information from several wavebands which can mitigate this problem and finally achieve clean aerial images. However, the distribution of values in an HSI vary drastically from that of RGB images. Thus, the reconstructed HSI cannot be directly used for mapping to RGB images.

Therefore, we train a ResNet-based architecture on the reconstructed HSI to obtain relevant feature distribution, which fits the distribution of clean dehazed RGB images. As an analogy to physical-chemical reactions, we refer to the obtained features as a Hyperspectral Catalyst. The HSC is attached with the 12-channel multi-cue input image as additional three channels to form a 15-channel input to cGAN. The ResNet uses the following loss $\Lagr_{r}$ for parameter optimization:
\begin{equation}
\label{eq7}
\begin{split}
\Lagr_{r} =& \Arrowvert y-G_r(G_h(G_x(x))) \Arrowvert 
\end{split}
\end{equation}
where $(x, y)$ refers to hazy RGB image and clean RGB image respectively. $G_r$ denotes the ResNet generator.

\begin{table*}[t]
    \centering
    \begin{tabular}{c|c c|c c|c c}
    \toprule
    \multirow{2}{*}{Method} & \multicolumn{2}{c|}{Thin Fog} & \multicolumn{2}{c|}{Moderate Fog} & \multicolumn{2}{c}{Thick Fog} \\ [5pt] \cline{2-3} \cline{4-5} \cline{6-7}
     & \multicolumn{1}{c}{SSIM} & \multicolumn{1}{c|}{PSNR} & \multicolumn{1}{c}{SSIM} & \multicolumn{1}{c|}{PSNR} & \multicolumn{1}{c}{SSIM} & \multicolumn{1}{c}{PSNR} \\ 
     \midrule
     \midrule
    Original & 0.7241  & 12.771  & 0.7399  & 12.587  & 0.4215  & 8.589  \\ 
        DCP~\cite{he2010single} & 0.7246  & 13.152  & 0.5735  & 9.783  & 0.5850  & 10.251  \\ 
        SAR-Opt-cGAN~\cite{grohnfeldt2018conditional} & 0.8419  & 20.195  & 0.7941  & 21.662  & 0.7573  & 19.655  \\ 
        Huang et al.~\cite{huang2020single} w/o SAR  & 0.8168  & 21.474  & 0.8274 & 22.095  & 0.7842 & 22.121  \\ 
        Huang et al.~\cite{huang2020single} SAR  & 0.9061  & 24.164  & \textbf{0.9264}  & 25.311  & 0.8640  & \textbf{25.073}  \\ 
        DehazeNet~\cite{cai2016dehazenet} & 0.8950  & 19.753  & 0.8552  & 18.125  & 0.7064  & 14.332  \\ 
        \textbf{SkyGAN} & \textbf{0.9248} & \textbf{25.381} & 0.9035 & \textbf{25.583} & \textbf{0.8925} & 23.430  \\
    \bottomrule
    \end{tabular}
    \caption{Comparative  results  of  the  proposed  method  and  existing dehazing  methods  over  SateHaze1k~\cite{huang2020single}}
    \label{table:haze1}
\end{table*}

\begin{figure*}[t!]
\centering
    \begin{subfigure}[t]{0.1\textwidth}
        \includegraphics[width=\textwidth]{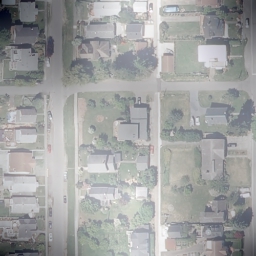}
        \includegraphics[width=\textwidth]{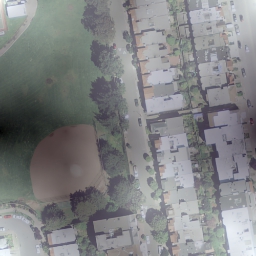}
        \caption{Hazy Input}
    \end{subfigure}
    \begin{subfigure}[t]{0.1\textwidth}
        \includegraphics[width=\textwidth]{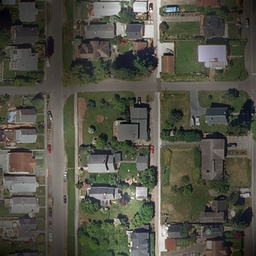}
        \includegraphics[width=\textwidth]{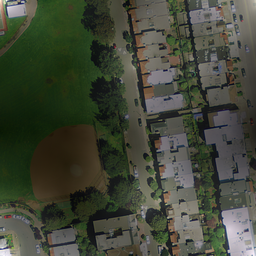}
        \caption{DCP~\cite{he2010single}}
    \end{subfigure}
    \begin{subfigure}[t]{0.1\textwidth}
        \includegraphics[width=\textwidth]{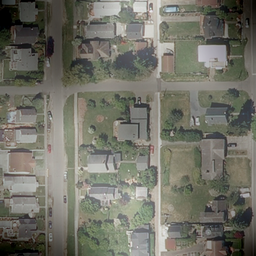}
        \includegraphics[width=\textwidth]{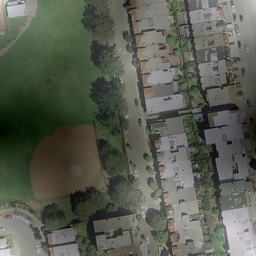}
        \caption{CAP~\cite{zhu2015fast}}
    \end{subfigure}
    \begin{subfigure}[t]{0.1\textwidth}
        \includegraphics[width=\textwidth]{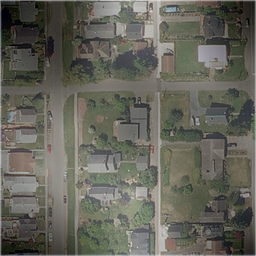}
        \includegraphics[width=\textwidth]{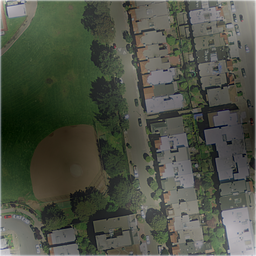}
        \caption{AOD~\cite{li2017aod}}
    \end{subfigure}
    \begin{subfigure}[t]{0.1\textwidth}
        \includegraphics[width=\textwidth]{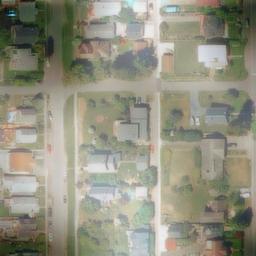}
        \includegraphics[width=\textwidth]{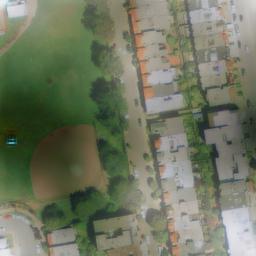}
        \caption{\footnotesize{CycleGAN}~\cite{zhu2017unpaired}}
    \end{subfigure}
    \begin{subfigure}[t]{0.1\textwidth}
        \includegraphics[width=\textwidth]{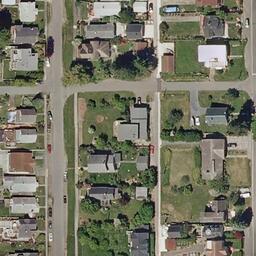}
        \includegraphics[width=\textwidth]{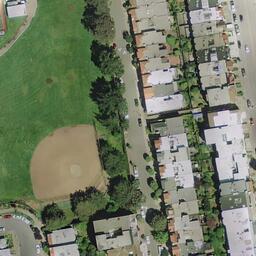}
        \caption{
        \footnotesize{GridDehazeNet}~\cite{liu2019griddehazenet}}
    \end{subfigure}
    \begin{subfigure}[t]{0.1\textwidth}
        \includegraphics[width=\textwidth]{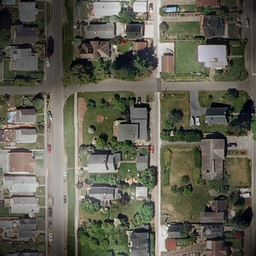}
        \includegraphics[width=\textwidth]{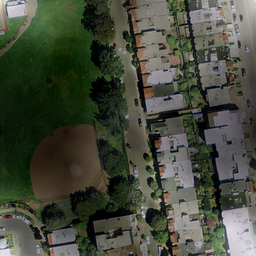}
        \caption{
        \footnotesize{GFN}~\cite{ren2018gated}}
    \end{subfigure}
    \begin{subfigure}[t]{0.1\textwidth}
        \includegraphics[width=\textwidth]{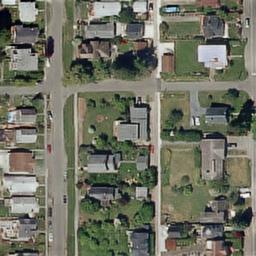}
        \includegraphics[width=\textwidth]{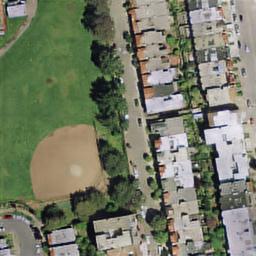}
        \caption{\textbf{SkyGAN}}
    \end{subfigure}
    \begin{subfigure}[t]{0.1\textwidth}
        \includegraphics[width=\textwidth]{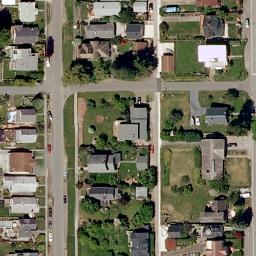}
        \includegraphics[width=\textwidth]{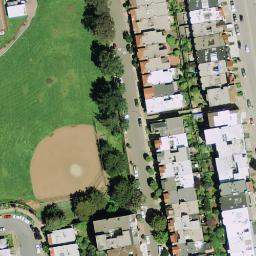}
        \caption{Ground Truth}
    \end{subfigure}
    \caption{Qualitative comparison on aerial images from HAI Dataset with a city-based scenario}
    \label{fig:city_qualitative}
    \medskip

\centering
    \begin{subfigure}[t]{0.1\textwidth}
        \includegraphics[width=\textwidth]{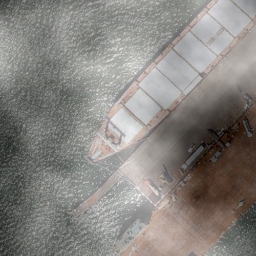}
        \includegraphics[width=\textwidth]{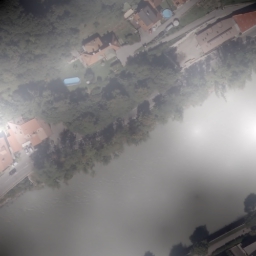}
        \caption{Hazy Input}
    \end{subfigure}
    \begin{subfigure}[t]{0.1\textwidth}
        \includegraphics[width=\textwidth]{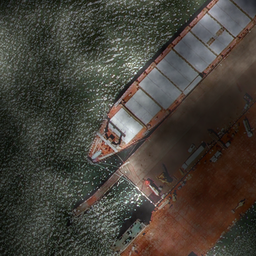}
        \includegraphics[width=\textwidth]{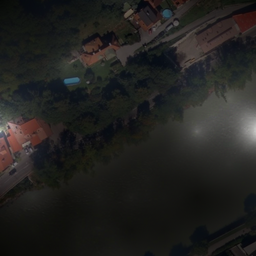}
        \caption{DCP~\cite{he2010single}}
    \end{subfigure}
    \begin{subfigure}[t]{0.1\textwidth}
        \includegraphics[width=\textwidth]{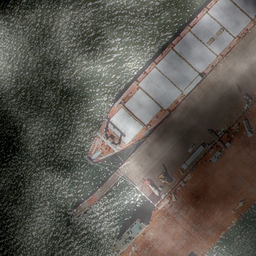}
        \includegraphics[width=\textwidth]{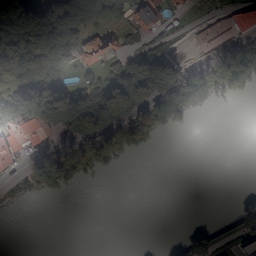}
        \caption{CAP~\cite{zhu2015fast}}
    \end{subfigure}
    \begin{subfigure}[t]{0.1\textwidth}
        \includegraphics[width=\textwidth]{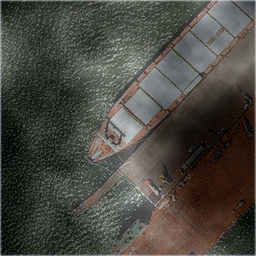}
        \includegraphics[width=\textwidth]{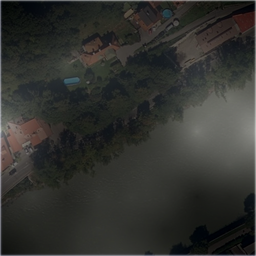}
        \caption{AOD~\cite{li2017aod}}
    \end{subfigure}
    \begin{subfigure}[t]{0.1\textwidth}
        \includegraphics[width=\textwidth]{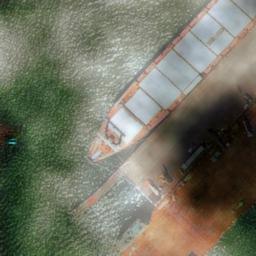}
        \includegraphics[width=\textwidth]{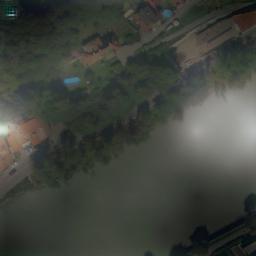}
        \caption{\footnotesize{CycleGAN}~\cite{zhu2017unpaired}}
    \end{subfigure}
    \begin{subfigure}[t]{0.1\textwidth}
        \includegraphics[width=\textwidth]{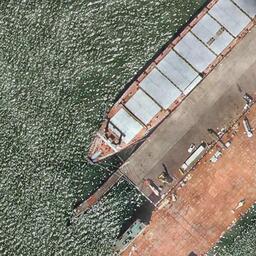}
        \includegraphics[width=\textwidth]{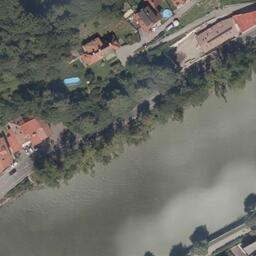}
        \caption{
        \footnotesize{GridDehazeNet}~\cite{liu2019griddehazenet}}
    \end{subfigure}
    \begin{subfigure}[t]{0.1\textwidth}
        \includegraphics[width=\textwidth]{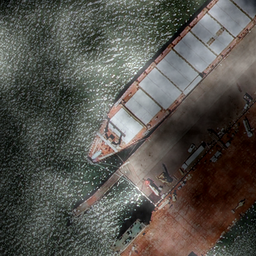}
        \includegraphics[width=\textwidth]{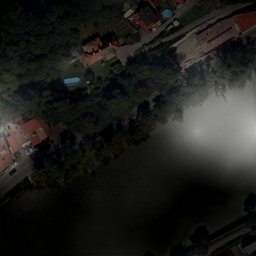}
        \caption{
        \footnotesize{GFN}~\cite{ren2018gated}}
    \end{subfigure}
    \begin{subfigure}[t]{0.1\textwidth}
        \includegraphics[width=\textwidth]{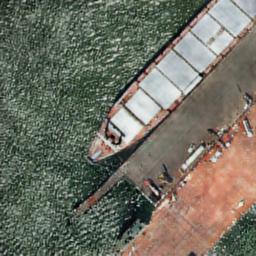}
        \includegraphics[width=\textwidth]{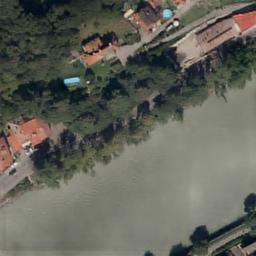}
        \caption{\textbf{SkyGAN}}
    \end{subfigure}
    \begin{subfigure}[t]{0.1\textwidth}
        \includegraphics[width=\textwidth]{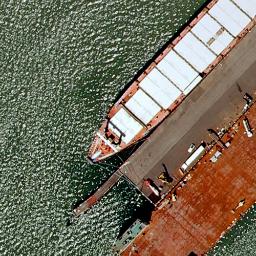}
        \includegraphics[width=\textwidth]{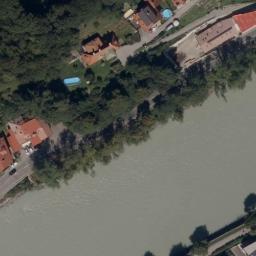}
        \caption{Ground Truth}
    \end{subfigure}
    \caption{Qualitative comparison on aerial images from HAI Dataset with water bodies and sky regions}
    \label{fig:water_qual}
    \medskip
%
    \begin{subfigure}[t]{0.1\textwidth}
        \includegraphics[width=\textwidth]{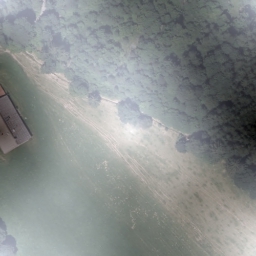}
        \includegraphics[width=\textwidth]{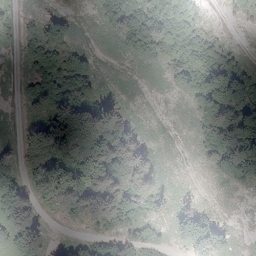}
        \caption{Hazy Input}
    \end{subfigure}
    \begin{subfigure}[t]{0.1\textwidth}
        \includegraphics[width=\textwidth]{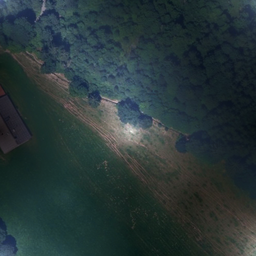}
        \includegraphics[width=\textwidth]{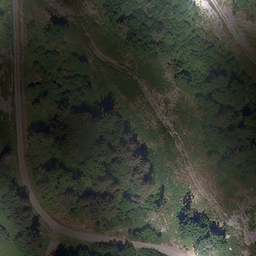}
        \caption{DCP~\cite{he2010single}}
    \end{subfigure}
    \begin{subfigure}[t]{0.1\textwidth}
        \includegraphics[width=\textwidth]{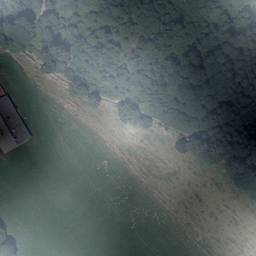}
        \includegraphics[width=\textwidth]{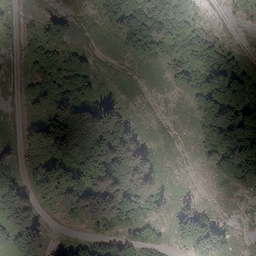}
        \caption{CAP~\cite{zhu2015fast}}
    \end{subfigure}
    \begin{subfigure}[t]{0.1\textwidth}
        \includegraphics[width=\textwidth]{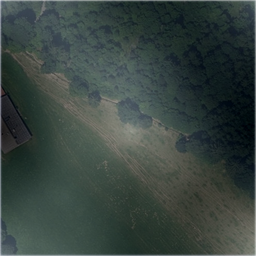}
        \includegraphics[width=\textwidth]{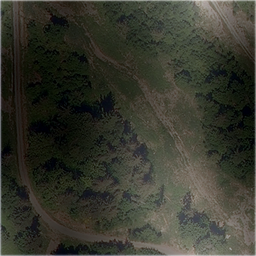}
        \caption{AOD~\cite{li2017aod}}
    \end{subfigure}
    \begin{subfigure}[t]{0.1\textwidth}
        \includegraphics[width=\textwidth]{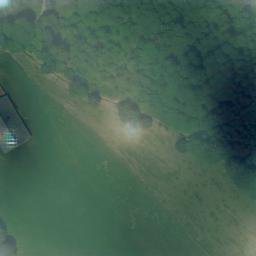}
        \includegraphics[width=\textwidth]{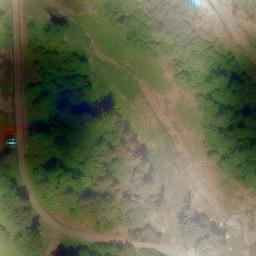}
        \caption{\footnotesize{CycleGAN}~\cite{zhu2017unpaired}}
    \end{subfigure}
    \begin{subfigure}[t]{0.1\textwidth}
        \includegraphics[width=\textwidth]{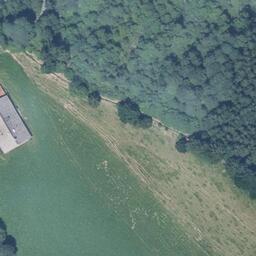}
        \includegraphics[width=\textwidth]{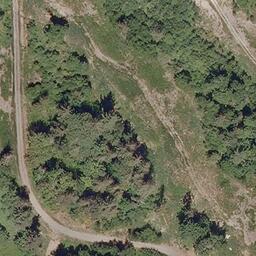}
        \caption{
        \footnotesize{GridDehazeNet}~\cite{liu2019griddehazenet}}
    \end{subfigure}
    \begin{subfigure}[t]{0.1\textwidth}
        \includegraphics[width=\textwidth]{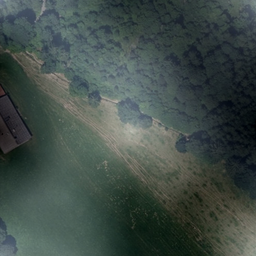}
        \includegraphics[width=\textwidth]{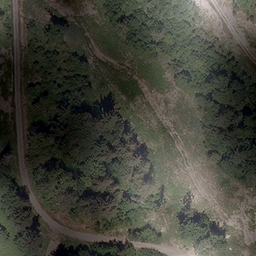}
        \caption{
        \footnotesize{GFN}~\cite{ren2018gated}}
    \end{subfigure}
    \begin{subfigure}[t]{0.1\textwidth}
        \includegraphics[width=\textwidth]{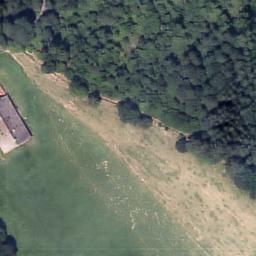}
        \includegraphics[width=\textwidth]{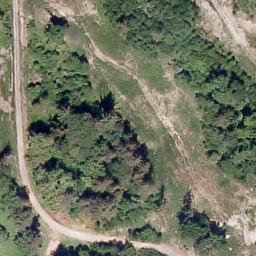}
        \caption{\textbf{SkyGAN}}
    \end{subfigure}
    \begin{subfigure}[t]{0.1\textwidth}
        \includegraphics[width=\textwidth]{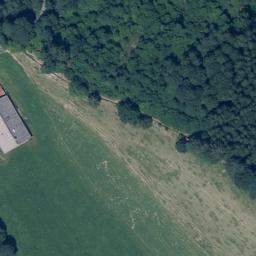}
        \includegraphics[width=\textwidth]{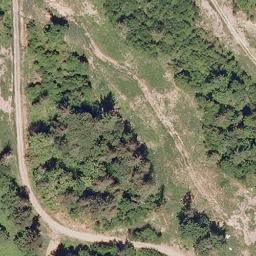}
        \caption{Ground Truth}
    \end{subfigure}
    \caption{Qualitative comparison on aerial images from HAI Dataset with substantial forest region}
    \label{fig:forest_qual}
\end{figure*}

\section{Dataset and Evaluation}\label{sec:data}
\paragraph{Hazy Aerial Image (HAI) Dataset} 
There has been tremendous development in the area of image restoration and enhancement, especially with the advent of deep learning. Standardized dataset benchmarks allow fair evaluation of different methodologies under similar evaluation criteria.  Several datasets such as Fattal's dataset~\cite{fattal2014dehazing}, FRIDA~\cite{tarel2012vision}, D-Hazy~\cite{ancuti2016dhazy}, and the Foggy Cityscapes datasets~\cite{sakaridis2018semantic} have been proposed for natural scene image dehazing. Apart from these, the RESIDE Dataset~\cite{reside2018} is a benchmark large-scale image dehazing for natural scene images. However, a standard large-scale benchmark for aerial images has been long expected. Recently, Huang et al.~\cite{huang2020single} proposed the SateHaze1k dataset, which consists of 1200 hazy aerial images. 

 The HAI Dataset is built upon the Inria Aerial Image Labelling (AIL) Dataset~\cite{maggiori2017can}. The AIL Dataset consists of 180 images each in train and test datasets. We crop these images using a sliding overlap with resolution 500 $\times$ 500, which results in $\sim$ 65,000 images in the training set. 
 The HAI dataset  is a large-scale dataset for fair evaluation and comparison for aerial image dehazing algorithms. The HAI dataset contains over 65 thousand pairs of synthetic hazy images, generated using 360 high-resolution aerial images. For every image, we create synthetic haze using diamond-square algorithm which generates a plasma fractal or the cloud fractal \cite{michaelis2019dragon}. The algorithm generates varying degree of vertical and horizontal perturbations. We implement five degrees of simulated haze using a varying grid size.
 Similarly, we create a test set of 600 images cropped from the AIL test partition.

To corroborate our results, we also evaluate on the recent SateHaze1k dataset~\cite{huang2020single}. The dataset contains 1200 pairs of hazy images. The images are divided into three levels of fog, namely thin, moderate, and thick fog.

\paragraph{CVPRW NTIRE \textquotesingle 18 and NTIRE \textquotesingle 20 datasets}
For HSI reconstruction task, we train our H2H module (Section \ref{subsec:h2h}) using NITRE 2018 (also known as BGU iCVL dataset) \cite{arad2016sparse,arad2018ntire} and NITRE 2020 hyperspectral datasets. The NTIRE 2018 dataset consists of 201 images, from both indoor and outdoor scenes. Along with RGB, they provide 31-channel HSI bands, separated by 10 nm for each image. Similarly, the NTIRE 2020 dataset consists a total of 360 images. For training purposes, the images were augmented by flipping and cropping randomly to generate a total of 6000 images.

\subsection{Quantitative Analysis}\label{subsec:quant}
For evaluating our model quantitatively, we use the test sets from the proposed HAI dataset and SateHaze1k~\cite{huang2020single} dataset. The results in terms of Peak Signal-to-Noise Ratio (PSNR) and Structural Similarity Index Measure (SSIM) are compiled in Table \ref{tab:my_label} and Table \ref{table:haze1}. The PSNR and SSIM metrics are computed between the de-hazed image and the ground truth clear image.

\begin{figure*}[t!]
\centering
    \begin{subfigure}[t]{0.135\textwidth}
        \includegraphics[width=\textwidth,height=20mm]{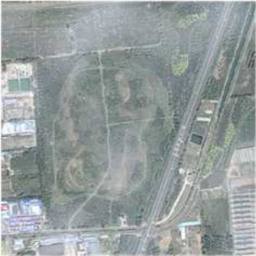}
        \includegraphics[width=\textwidth,height=20mm]{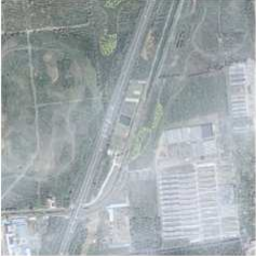}
        \includegraphics[width=\textwidth,height=20mm]{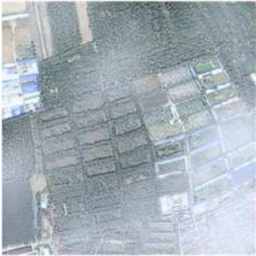}
        \caption{Hazy Input}
    \end{subfigure}
    \begin{subfigure}[t]{0.135\textwidth}
        \includegraphics[width=\textwidth,height=20mm]{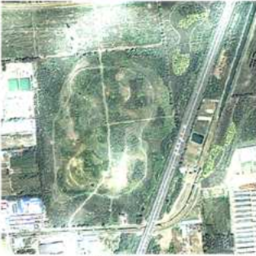}
        \includegraphics[width=\textwidth,height=20mm]{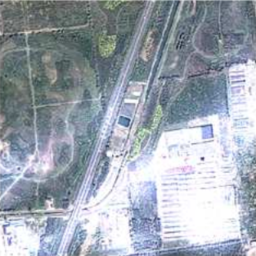}
        \includegraphics[width=\textwidth,height=20mm]{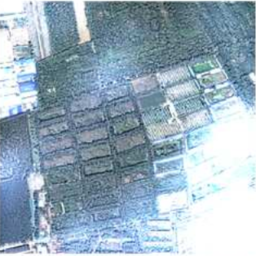}
        \caption{DCP~\cite{he2010single}}
    \end{subfigure}
    \begin{subfigure}[t]{0.135\textwidth}
        \includegraphics[width=\textwidth,height=20mm]{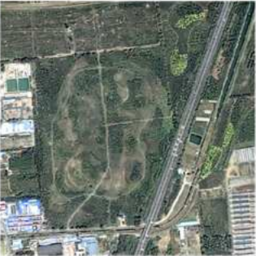}
        \includegraphics[width=\textwidth,height=20mm]{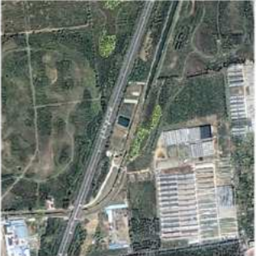}
        \includegraphics[width=\textwidth,height=20mm]{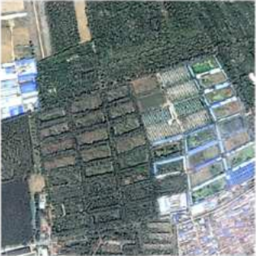}
        \caption{\footnotesize Huang et al.\ ~\cite{huang2020single}}
    \end{subfigure}
    \begin{subfigure}[t]{0.135\textwidth}
        \includegraphics[width=\textwidth,height=20mm]{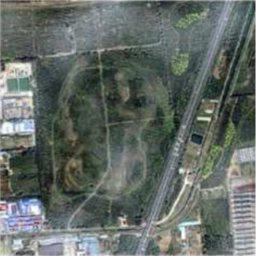}
        \includegraphics[width=\textwidth,height=20mm]{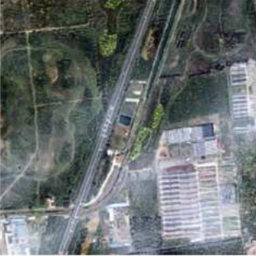}
        \includegraphics[width=\textwidth,height=20mm]{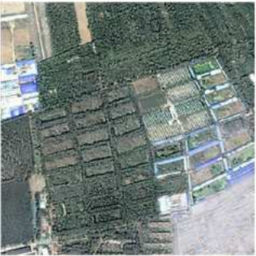}
        \caption{DehazeNet~\cite{cai2016dehazenet}}
    \end{subfigure}
    \begin{subfigure}[t]{0.135\textwidth}
        \includegraphics[width=\textwidth,height=20mm]{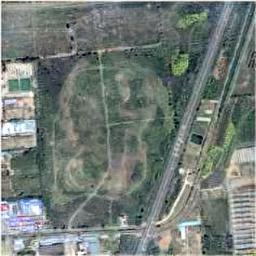}
        \includegraphics[width=\textwidth,height=20mm]{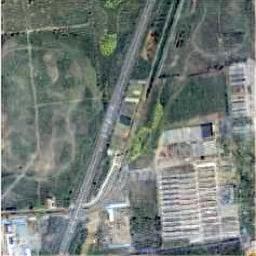}
        \includegraphics[width=\textwidth,height=20mm]{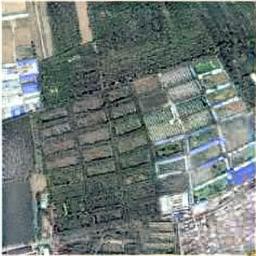}
        \caption{\textbf{SkyGAN}}
    \end{subfigure}
    \begin{subfigure}[t]{0.135\textwidth}
        \includegraphics[width=\textwidth,height=20mm]{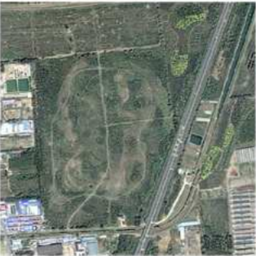}
        \includegraphics[width=\textwidth,height=20mm]{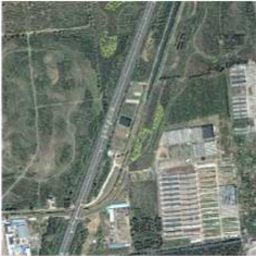}
        \includegraphics[width=\textwidth,height=20mm]{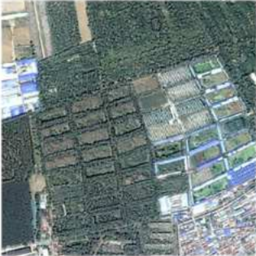}
        \caption{Ground Truth}
    \end{subfigure}
\caption{Qualitative comparison of SkyGAN with state-of-the-art techniques over SateHaze1k~\cite{huang2020single} dataset}
\label{fig:SAR_qual}
\end{figure*}

\paragraph{Performance on HAI Dataset}
 To have a fair comparison of the model on the proposed dataset, we evaluate several prior-based and pre-trained deep learning-based models on HAI dataset. For prior-based techniques, we use the Dark Channel Prior (DCP)~\cite{he2010single} and Color Attenuation Prior (CAP)~\cite{zhu2015fast}. 
For deep learning-based techniques, we take into account the state-of-the-art AOD-Net~\cite{li2017aod},  CycleGAN~\cite{zhu2017unpaired}, 
GFN \cite{ren2018gated} and GridDehazeNet \cite{liu2019griddehazenet}.
The results are presented in Table \ref{tab:my_label}, which  indicates the percentage increase SkyGAN could achieve over rest of the techniques. It is quite evident from the metrics that the proposed SkyGAN performs extremely well in comparison to the rest, both in terms of PSNR and SSIM.

\paragraph{Performance on SateHaze1k Dataset.}
SateHaze1k consists of 45 images from each of the three levels of fog  (thin, moderate, and thick). Table \ref{table:haze1} shows the quantitative comparison between various techniques and SkyGAN. SkyGAN obtains the highest SSIM in thin and thick fog scenarios. Similarly, our method achieves the best PSNR in thin and moderate fog conditions. Overall, the proposed method outperforms the existing state-of-the-art \cite{huang2020single} in 4 out of 6 performance measures (SSIM and PSNR) across the three types of fog conditions.

\subsection{Qualitative Analysis}\label{subsec:qualitative}
The visual analysis for multiple aerial hazy images was carried out from the HAI and SateHaze1k datasets. From the HAI, a set of six challenging aerial hazy images and the recovered clear images using the proposed SkyGAN and existing methods are compared. To show the robustness of our method, we considered various scenarios from urban regions (in Figure \ref{fig:city_qualitative}), water bodies and sky with large white/blue regions (in Figure \ref{fig:water_qual}), and forest regions (in Figure \ref{fig:forest_qual}). It can be easily assessed that existing methods are able to reduce haze in some regions but fail on certain areas such as grass fields (in Figure \ref{fig:city_qualitative}), red colored ship decks (in Figure \ref{fig:water_qual}), and large dense forests (in Figure \ref{fig:forest_qual}). In contrast, the proposed SkyGAN removes the haze and also restores the color balance in the recovered clear image.\par

Furthermore, we compare the qualitative results of the proposed and existing methods for three aerial hazy images from SateHaze1k. The images are captured from agricultural fields, roads, residential areas and empty grounds (Figure \ref{fig:SAR_qual}). The proposed method is reducing the haze in the entire image as opposed to the non-uniform improvements obtained by the existing methods. In terms of closeness to the actual ground truth, our results are better than the existing state-of-the-art \cite{huang2020single}. The work \cite{huang2020single} introduces darker colors in the image as opposed to the actual light colors in the clear images. The higher quantitative performance (PSNR and SSIM) of SkyGAN further proves its robustness to multiple scenarios.

\begin{figure*}
    \centering
    \begin{subfigure}[t]{0.13\textwidth}
    \caption{Hazy Image}
    \includegraphics[width=\textwidth,height=22mm]{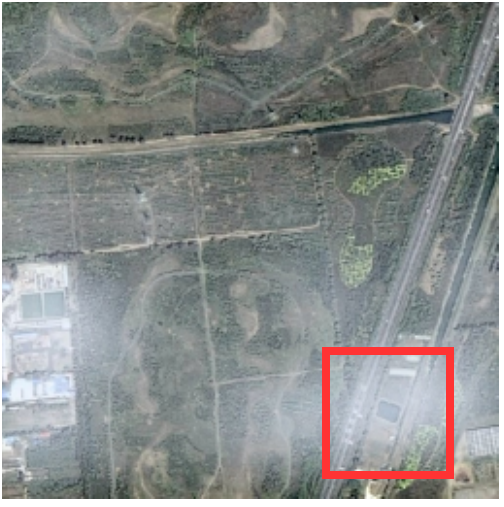}
    \end{subfigure}
    \begin{subfigure}[t]{0.13\textwidth}
    \caption{Model-1}
    \includegraphics[width=\textwidth,height=22mm]{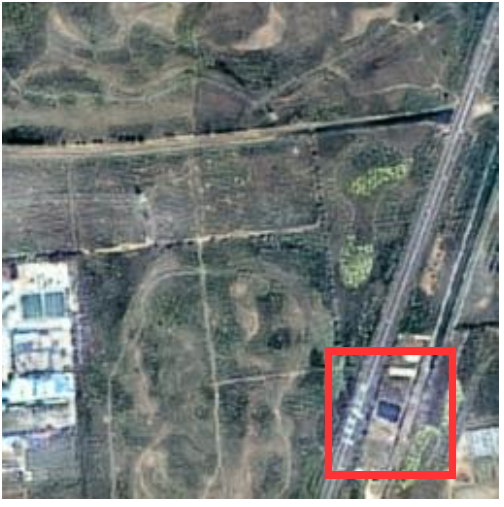}
    \end{subfigure}
    \begin{subfigure}[t]{0.13\textwidth}
    \caption{Model-2}
    \includegraphics[width=\textwidth,height=22mm]{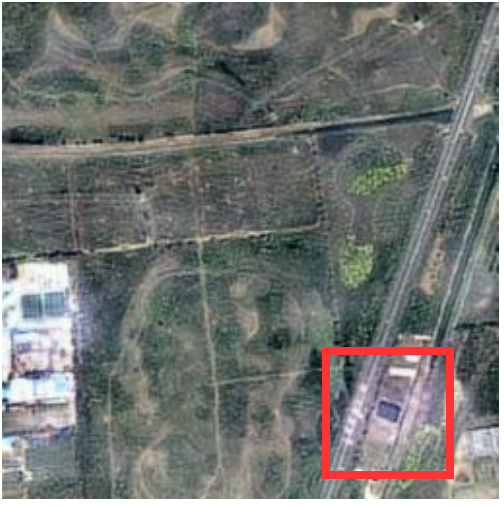}
    \end{subfigure}
    \begin{subfigure}[t]{0.13\textwidth}
    \caption{Model-3}
    \includegraphics[width=\textwidth,height=22mm]{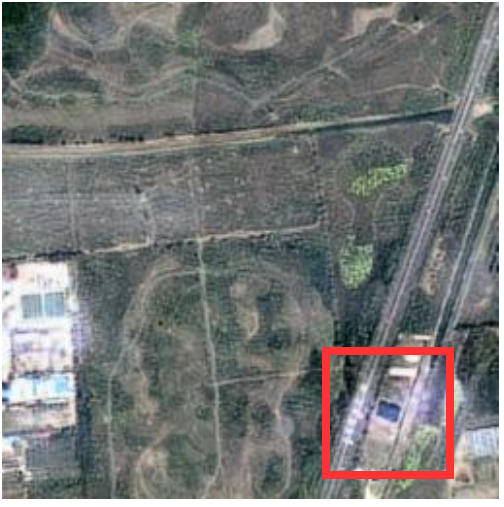}
    \end{subfigure}
    \begin{subfigure}[t]{0.13\textwidth}
    \caption{Model-4}
    \includegraphics[width=\textwidth,height=22mm]{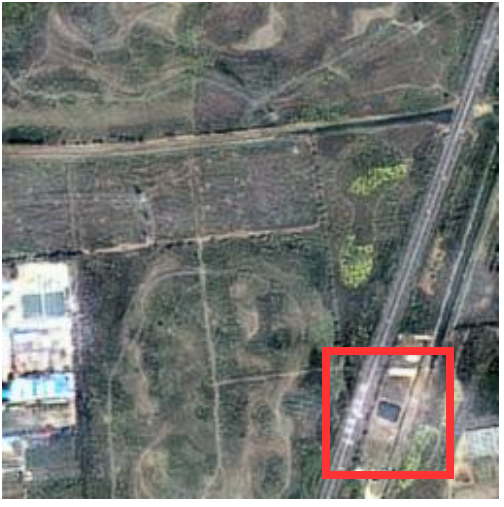}
    \end{subfigure}
    \begin{subfigure}[t]{0.13\textwidth}
    \caption{\textbf{SkyGAN}}
    \includegraphics[width=\textwidth,height=22mm]{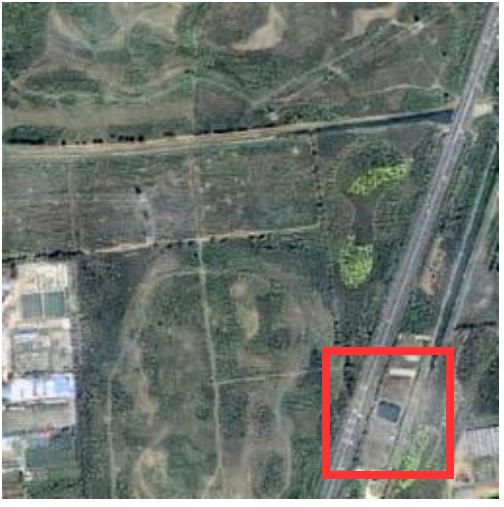}
    \end{subfigure}
    \begin{subfigure}[t]{0.13\textwidth}
    \caption{Ground Truth}
    \includegraphics[width=\textwidth,height=22mm]{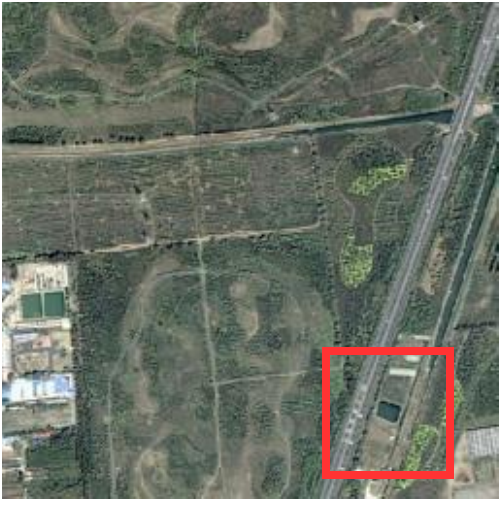}
    \end{subfigure}
    
    \caption{Qualitative comparison for ablation models as described in Table \ref{table:ablation_study}. The gradual improvement in visual as highlighted by the red box, clearly indicate the additive enhancement of each component.} 
    \label{fig:ablation_comparison}
\end{figure*}

\section{Ablation Study}\label{sec:ablation}
\begin{figure}
    \centering
    \includegraphics[width=\linewidth]{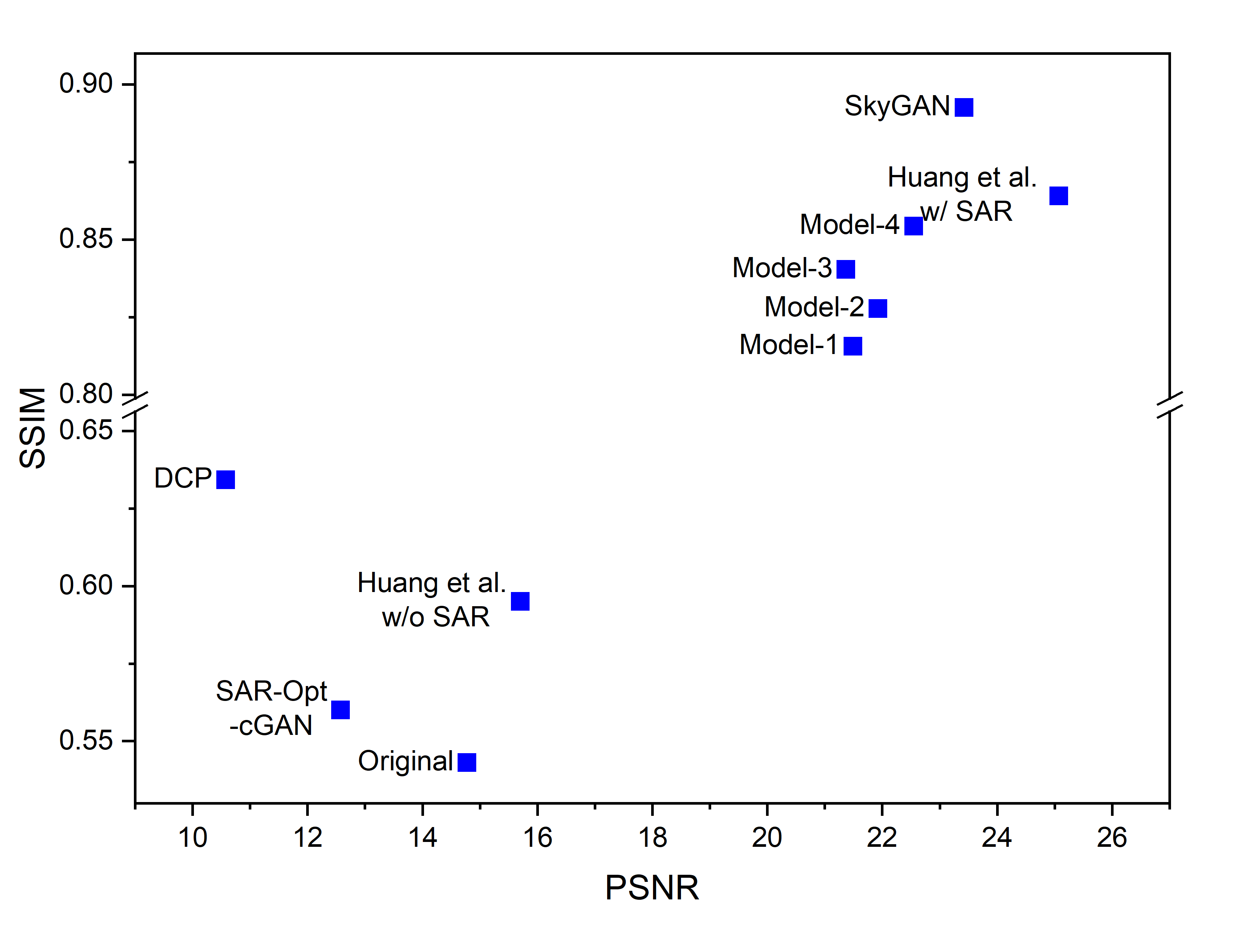}
    \caption{
    Quantitative comparison for ablation models along with existing state-of-the-art techniques.}
    \label{fig:ablation-quant}
\end{figure}

\begin{table}
\begin{center}
\resizebox{\columnwidth}{!}{
    \begin{tabular}{c|c c c c}
    \toprule
    &  \multicolumn{4}{c}{Components} \\ [4pt]
    \cline{2-5}
    Method & \begin{tabular}[c]{@{}c@{}}Hyperspectral \\Catalyst\end{tabular} & \begin{tabular}[c]{@{}c@{}}Task\\ Supervision\end{tabular} & \begin{tabular}[c]{@{}c@{}}Domain\\Classifier \end{tabular} & \begin{tabular}[c]{@{}c@{}}Multi-cue\\ Color Space\end{tabular} \\ \midrule
    Model-1 & & & &  \\
    Model-2 & \checkmark & & &  \\
    Model-3 & \checkmark & \checkmark & &  \\
    Model-4 & \checkmark & \checkmark &\checkmark & \\
    \textbf{SkyGAN} & \checkmark & \checkmark & \checkmark & \checkmark \\
    \bottomrule
    \end{tabular}}
\end{center}
\caption{Component-wise ablation models}
\label{table:ablation_study}
\end{table}
In this subsection, we demonstrate the effectiveness of various components that have successfully contributed towards dehazing the aerial images through a detailed ablation study. This experiment includes 4 progressive models and the SkyGAN as shown in Table \ref{table:ablation_study}. We add the various components one-by-one and evaluate their respective performance. We report results on the thick fog level of SateHaze1k dataset for the following ablation models:

\textbf{Backbone Architecture}: The choice of the right image translation model is crucial for the quality of output dehazed images. As conditional GANs (cGAN) has been widely accepted in literature for a variety of tasks, we choose a modified cGAN as the baseline for our experiments, referred to as Model-1. The successive models differ significantly from Model-1 as it is exclusively based on RGB imformation. On the other hand, Model 2-4 incorporate HSI information for dehazing. Figure \ref{fig:ablation_comparison} and Figure \ref{fig:ablation-quant} shows the performance of proposed SkyGAN in comparison to different ablation models along with different representative state-of-the-art techniques for image dehazing respectively.\par

\textbf{Hyperspectral Catalyst, Task Supervision, Domain Classifier}: To investigate the utility of these auxiliary networks for aerial-image dehazing, we train Model-1 by adding the components progressively. Figure \ref{fig:ablation-quant} depicts the performance of various models. Quantitatively, these components leads to increment in performance by 1.5\% in terms of SSIM. This can be confirmed by comparing the visual results as shown in Figure \ref{fig:ablation_comparison}.\par
\textbf{Multi-cue Color Space}:  
In contrast to other components, adding multi-cue color spaces leads to 4.5\% improvement in SSIM. This is depicted in Figure \ref{fig:ablation-quant} as Model 1-4 are clustered in an area, and SkyGAN stands ahead of all these ablation model. Thus, multi-cue color space forms a key contributor towards the performance of SkyGAN.


\section{Conclusions}\label{sec:conclusion}
The paper presents a novel GAN based framework, SkyGAN, which utilizes HSI guidance and multi-cue color input for aerial image dehazing. SkyGAN effectively combines adversarial distribution discrepancy alignment and cycle-consistency constraint to reconstruct domain-aware HSI from a set of hyperspectral natural scene images in an unpaired manner. A residual network is further used to alleviate data distribution discrepancy from reconstructed HSI, thereby generating a HSC. The HSC along with a 12-channel multi-cue input is used by the cGAN module (I2I) to perform effective dehazing. The hyperspectral catalyst consists of rich content from several visual bands, which is coupled with the multi-cue color space, leading to highly effective aerial image dehazing. In addition, we also propose a large-scale dataset (HAI) for haze removal in aerial images consisting of realistic, non-homogeneous haze with varying density, along with an extensive evaluation of the dataset on a representative set of existing state-of-the-art techniques. 

\section*{Acknowledgements}
This work is supported by BITS Additional Competitive Research Grant (PLN/AD/2018-19/5).

{\small
\bibliographystyle{ieee_fullname}
\bibliography{udodai}
}

\end{document}